%% file: main.tex
\documentclass[letterpaper, 10 pt, journal, twoside]{ieeetran}

\IEEEoverridecommandlockouts                              

\usepackage{graphicx}
\usepackage{bbold}
\usepackage{mathtools}

\usepackage{float} 
\usepackage{amsmath} 
\usepackage{amssymb}
\usepackage{amsfonts}
\usepackage{amstext}

\usepackage{amsthm}
\usepackage[T1]{fontenc}
\usepackage[symbolgreek,italic]{mathastext}

\usepackage{bbm}

\usepackage{algorithm}
\usepackage[noend]{algpseudocode}

\let\labelindent\relax
\usepackage{enumitem}

\usepackage{booktabs}
\usepackage{multirow}
\usepackage{stfloats}
\usepackage{array}

\usepackage{fancyhdr}
\fancypagestyle{plain}{
  \fancyhf{}
  \fancyhead[C]{Conference on \LaTeX}     
  \fancyfoot[L]{This is a notice}

}
\usepackage{eso-pic}
\usepackage{hyperref}

\AddToShipoutPictureBG*{%
  \AtPageLowerLeft{%
    \setlength\unitlength{1in}%
    \hspace*{\dimexpr0.5\paperwidth\relax}
    \makebox(0,0.75)[c]{\parbox{0.7\paperwidth}{\raggedright\footnotesize© 2021 IEEE.  Personal use of this material is permitted. Permission from IEEE must be obtained for all other uses, in any current or future media, including reprinting/republishing this material for advertising or promotional purposes, creating new collective works, for resale or redistribution to servers or lists, or reuse of any copyrighted component of this work in other works.}}%
}}

\title{Informing Real-time Corrections in Corrective Shared Autonomy Through Expert Demonstrations}

\author{Michael Hagenow,$^{1}$ Emmanuel Senft,$^{2}$ Robert Radwin,$^{3}$ Michael Gleicher,$^{2}$ Bilge Mutlu,$^{2}$ Michael Zinn$^{1}$
\thanks{Manuscript received: February, 24, 2021; Revised May, 21, 2021; Accepted June, 16, 2021.}
\thanks{This paper was recommended for publication by Editor Jee-Hwan Ryu upon evaluation of the Associate Editor and Reviewers' comments. This work was supported by a NASA University Leadership Initiative (ULI) grant awarded to the UW-Madison and The Boeing Company (Cooperative Agreement \# 80NSSC19M0124).}
\thanks{$^{1}$Michael Hagenow and Michael Zinn are with the Department of Mechanical
Engineering, University of Wisconsin--Madison, Madison 53706, USA
        {\tt\small [mhagenow|mzinn]@wisc.edu}}%
\thanks{$^{2}$Emmanuel Senft, Michael Gleicher, and Bilge Mutlu are with the Department of Computer
Sciences, University of Wisconsin--Madison, Madison 53706, USA
        {\tt\small [esenft|gleicher|bilge]@cs.wisc.edu}}
\thanks{$^{3}$Robert Radwin is with the Department of Industrial and Systems Engineering, University of Wisconsin--Madison, Madison 53706, USA
        {\tt\small rradwin@wisc.edu}}
\thanks{Digital Object Identifier (DOI): \href{https://doi.org/10.1109/LRA.2021.3094480}{10.1109/LRA.2021.3094480}.}}%

\begin{document}
\maketitle
\markboth{IEEE Robotics and Automation Letters. Preprint Version. Accepted June, 2021}
{Hagenow \MakeLowercase{\textit{et al.}}: Informing Real-time Corrections in Corrective Shared Autonomy Through Expert Demonstrations} 

\input{abstract.tex}
\input{introduction}
\input{relatedwork}
\input{method}
\input{implementation}
\input{experimentalevaluation}

\input{conclusions}

\bibliographystyle{IEEEtran}
\bibliography{references.bib}
\end{document}

%% file: abstract.tex
\begin{abstract}
\emph{Corrective Shared Autonomy} is a method where human corrections are layered on top of an otherwise autonomous robot behavior. Specifically, a Corrective Shared Autonomy system leverages an external controller to allow corrections across a range of task variables (e.g., spinning speed of a tool, applied force, path) to address the specific needs of a task. However, this inherent flexibility makes the choice of what corrections to allow at any given instant difficult to determine. This choice of corrections includes determining appropriate robot state variables, scaling for these variables, and a way to allow a user to specify the corrections in an intuitive manner. This paper enables efficient Corrective Shared Autonomy by providing an automated solution based on Learning from Demonstration to both extract the nominal behavior and address these core problems. Our evaluation shows that this solution enables users to successfully complete a surface cleaning task, identifies different strategies users employed in applying corrections, and points to future improvements for our solution.
\end{abstract}

%% file: introduction.tex
\section{INTRODUCTION}
\begin{figure}[t]
\centering
\includegraphics[width=3.4in]{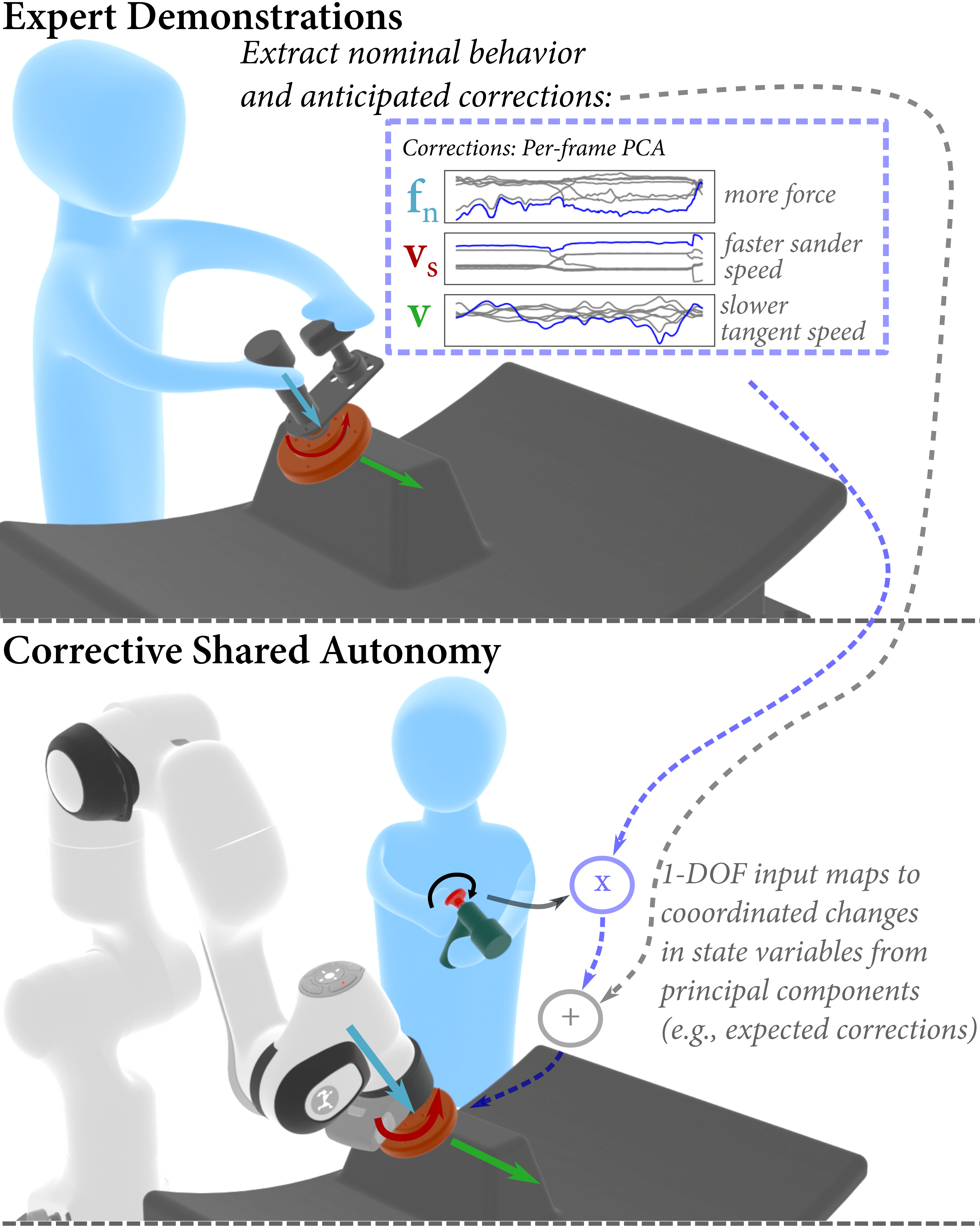}
\caption{In this paper, we present a method to automatically infer the nominal behavior and types of corrections needed in \emph{Corrective Shared Autonomy}. \textit{Top:} An expert sander provides demonstrations that are used to learn the nominal behavior and the most likely correction: a coordination of tangential speed, tool speed, and applied force. \textit{Bottom:} The operator uses a 1-DOF input to provide simultaneous corrections to these three robot state variables.}
\label{fig:teaser}
\vspace{-15pt}
\end{figure}
\IEEEPARstart{O}{ne} key challenge in deploying robotic solutions is dealing with uncertainty and variability in complex tasks, such as those involving physical interaction. When complete automation is not possible, a potential solution is to blend together human input and an autonomous behavior into a shared autonomy system. One paradigm for situations where the task has variability, but also has a structured task model (e.g., making sequential passes to paint a surface) is to allow human operators to provide corrections to otherwise autonomous robot behavior to address issues on an as-needed basis. As one example, the operator may augment the position and orientation of the robot's end effector for a misaligned bolt insertion. As another example in robot sanding, the operator may augment the applied force, sanding speed, and angle of attack of the robot to address a tough area during composite sanding. The diversity of variables which may require corrections for any given task points towards a solution where a central interface allows corrections across a host of robot state variables. In previous work \cite{HagenowCSA2021}, we introduced the \emph{Corrective Shared Autonomy} framework where users provide corrective input through an external device to any relevant robot state variable, dictated by the needs of a given task. For example, this method allowed end-users to provide input using a 3-degree-of-freedom (DOF) haptic device which applied corrections (i.e., differential changes) to robot state variables such as applied force, execution speed, and the surface path to address unexpected pockets and bubbles of air during a mock-up composite layup task. While this work showed that \emph{Corrective Shared Autonomy} is possible where the nominal behavior and allowable corrections were manually determined for tasks (e.g., manually parameterized passes over the surface and a set magnitude for changes to the applied force), the core challenges of how to choose appropriate correction variables and magnitudes have not yet been addressed. In this paper, we focus on the preliminary step of determining the corrections a user can make during the task execution. We provide a method to leverage expert demonstrations to automatically determine a nominal behavior, correction variables, scaling, and input mapping for a \emph{Corrective Shared Autonomy} system.

One common method for designing robot task models is to use Learning from Demonstration (LfD) \cite{ArgallLFD2009}. In this paradigm, experts can provide demonstrations through various modalities (e.g., natural demonstration, kinesthetic teaching) from which techniques such as direct and indirect learning can determine suitable robot behaviors. Our premise is that LfD is a logical way to inform both the nominal robot behavior and the range of corrections that might be expected during robot execution of the task. Previous methods in shared control leverage variance in LfD to inform input blending and haptic guidance during teleoperation. For \emph{Corrective Shared Autonomy}, the variance seen between demonstrations can serve as an indicator of corrections that may be necessary. Specifically, by looking at the principal components of variation over the course of the demonstrations, we can extract relevant task variables, including coordinations of variables, that may need corrections. For example, in Figure \ref{fig:teaser}, coordinated variability in applied force and sanding speed during demonstrations indicates that similar coordinated corrections may be needed depending on the surface hardness of the piece. The operator may also have to make larger corrections than what appeared in the expert demonstrations, which can be addressed by an override strategy. We focus on the types of corrections commonly seen in demonstrations to make it feasible to map corrections through a limited degree-of-freedom input.

Even with information about the corrections required for a given task, choosing how a human can provide these corrections involves addressing a number of challenges. When the corrections are kinematic (e.g., end-effector position) and the input is spatially grounded (e.g., fixed 3-DOF input), there is an obvious choice for how corrective input should map to the robot. As corrections generalize to variables without a clear spatial correlation (e.g., sanding speed, temperature) or when the input device itself is not spatially grounded (e.g., mobile input), it is less clear how these corrections can be made effectively. To address these challenges, we propose heuristics for mapping generalized corrective input to both a mobile 1-DOF input and a 3-DOF grounded input.

In this paper, we present a method that leverages Learning from Demonstration to determine the nominal behavior and correction variables, scaling, and mapping for a \emph{Corrective Shared Autonomy} system. Our contributions include:

\begin{enumerate}[leftmargin=*]
    \item Describing the design space and considerations related to correction variables, scaling, and input mapping
    \item Introducing methods for \emph{Corrective Shared Autonomy} to extract correction scaling and variables using Principal Component Analysis and to map the results to both 1-DOF and 3-DOF haptic interfaces
    \item Implementing an override strategy combining proportional and integral input which allows corrective input both within and outside the ranges seen in demonstrations
    \item User testing of the 1-DOF and 3-DOF input devices in surface cleaning tasks situated in aircraft manufacturing
\end{enumerate}

%% file: relatedwork.tex
\section{RELATED WORK}
\label{sec:relatedwork}
In this work, we propose a method to learn the nominal behavior and correction scaling, variables, and input mapping in \emph{Corrective Shared Autonomy} via Learning from Demonstration. To contextualize our contributions, we provide a brief review of the use of Learning from Demonstration in shared autonomy as well as a summary of how learning is incorporated in other methods for real-time robot corrections.

Learning from Demonstration has been used previously in shared autonomy, however, the focus has been to inform shared control policies during teleoperation. Pérez-del-Pulgar et al. \cite{PerezdelPulgar2016} record end-effector forces and torques while kinesthetically teaching a robot arm a peg-in-hole sequence. The demonstrations are encoded via Gaussian Mixture Regression (GMR)s to provide haptic guidance forces towards the expert strategy during later teleoperation of the task. Abi-Farraj et al. \cite{Giordano2017} use teleoperation demonstrations of grasping a desired object to encode a shared control policy over conditional trajectory distributions
to provide haptic guidance during robot teleoperation grasping tasks. Finally, Zeestraten et al. \cite{CalinonRAL2018} use demonstrations of a cap-replacement task to encode shared control policies using a Gaussian mixture model (GMM) on a Riemannian manifold and compare teleoperation between haptic and state shared control.

Previous methods for providing real-time corrections have used corrections to refine robot behaviors, but have not used learning to inform the types of corrections that may be necessary during a task. Many methods in physical human-robot interaction (pHRI) \cite{LoseyDeformations2018}\cite{Ude2018}\cite{Liu2017} have been proposed to refine robot trajectories or policies based on physical forces exerted on a robot. For example, in Bajcsy et al. \cite{LoseyPolicy2017} and Losey et al. \cite{LoseyPHRI2020}, physical interventions on robot joints are used to successfully infer parameters of an optimal policy (e.g., restrictions on orientation). In each of these works, emphasis is placed on determining a change to the original behavior or policy, which can be  successively refined. These works either do not leverage prior task knowledge to inform necessary corrections or do so in a very limited way, for example dictating the stiffness of the robot impedance controller based on the covariance of the expert demonstrations \cite{Ude2018}.

The approach presented in this paper focuses on when the corrections are generalized beyond kinematic variables and inputted using an external controller. Existing methods for corrections using an external controller \cite{Masone2014,cognetti2020,srinivasa2020} allow for path transformations and policy learning of UAVs and mobile robots. However, similar to the methods in pHRI, the main emphasis is on corrections to kinematics and on updating the trajectory or policy rather than using task knowledge to inform possible corrective input. In \emph{Corrective Shared Autonomy}, identifying what corrections a task could require is crucially important to address the flexibility of corrections, both in terms of robot state variables and magnitude. Developing methods to learn needed corrections is a crucial step towards allowing for generalized corrections across a range of complex tasks.

%% file: method.tex
\begin{figure*}
\centering
\includegraphics[width=.95\textwidth]{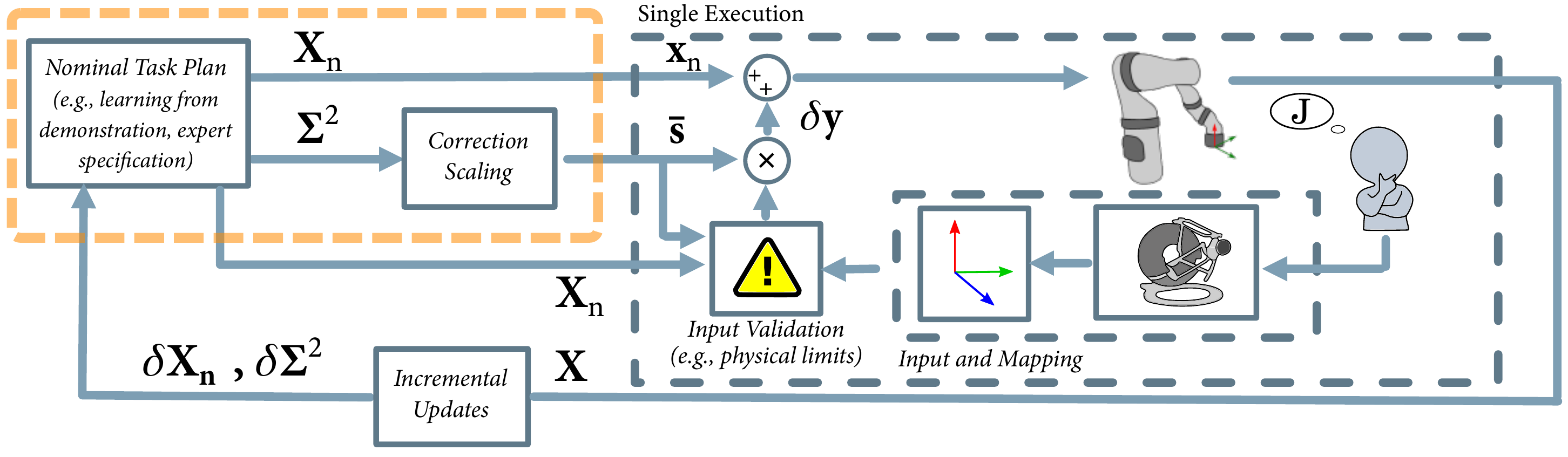}
\caption{Topology of the \emph{Corrective Shared Autonomy} system. In this work, we focus on generating nominal behaviors ($\textbf{X}_n$) and correction scaling ($\bar{\textbf{s}}$) from variance ($\mathbf{\Sigma}^{2}$) in demonstrations (orange box). During execution, the user makes corrections to nominal state, $\textbf{x}_n$, based on an internal, expert cost function, $J$, and provides corrections in a subspace of the state vector. After, this new execution example, $\textbf{X}$, can be used to update the models.}
\label{fig:correctionsblockdiagram}
 \vspace{-5pt}
\end{figure*}

\section{METHOD AND CONSIDERATIONS}\label{section:method}
The arbitration of \emph{Corrective Shared Autonomy} \cite{HagenowCSA2021} can be expressed by summing autonomous input from a nominal task model and relative human input based on observations of problems caused by the current robot state (see Figure \ref{fig:correctionsblockdiagram}). The final commanded robot state, $\textbf{x}$, can be calculated as:
\begin{align}
        \textbf{x} = \textbf{x}_{n} + \delta \textbf{y},\textbf{x}_{n} \in \mathbb{R}^{m}, \delta \textbf{y} \in S(\mathbb{R}^{m})
\label{eq:arbitration}
\end{align}
where $\textbf{x}_{n}$ is the nominal robot state, $\delta \textbf{y}$ is the correction to the robot state variables, $m$ is the dimension of the robot state, and $S(\mathbb{R}^{m})$ is the subspace where users can provide corrections. The correction can be further decomposed into:
\begin{equation}
    \delta \textbf{y} = \textbf{f}(\textbf{u})
\end{equation}
where $\textbf{u}$ is the user input and $\textbf{f}$ is a function that scales and maps the user input to corrections in the robot space. In this section, we introduce the design space for selecting the nominal behavior, correction variables and scaling (i.e., types of corrections), and a mapping for the human input.

\subsection{Extracting Correction Scaling and Variables}
Allowing for arbitrary robot state corrections requires a method to determine what corrections can be given at any point in the execution. For example, a correction to force may be needed during a delicate surface interaction whereas corrections to position may be needed later for interacting with a work piece that is not positioned properly. While it is possible for a designer to choose the types of corrections for a particular task, automatic inference could simplify adoption for new tasks. There may be many ways to infer the types of corrections needed for a given task. One option is to derive the nominal behavior from the mean behavior of task executions and necessary correction scaling and variables can be inferred from an analysis of the differences between executions of a given task. For example, potential corrections could come from variance in expert demonstration. Regardless, there are a few main considerations for any corrective method when determining the scaling and variables:
\begin{itemize}[leftmargin=*]
  \item \textbf{Consistency}: Necessary corrections can vary over the task execution. It may be desirable for the allowable corrections (variables and magnitude) to adapt to match the needs of the task at any given moment, however, it may also be desirable to apply filtering so that the corrections linked to user inputs are mostly consistent (i.e., change slowly). If consistency is crucial, such as when the user needs to make the input without seeing how the system reacts, it may be desirable to use a technique such as a piece-wise constant approximation \cite{binev2005universal}. However, such techniques introduce challenges (e.g., transitions in behavior).
  \item \textbf{Saturation}: There may be instances where the corrections exceed what can be mapped to the robot. For example, a correction may induce a movement outside of the robot's reach or a force outside the robot's manipulability. In these cases, it must be considered whether to saturate the user's command or to modify the task model to a behavior suitable for the robot platform (e.g., change the orientation of the workpiece so the robot can apply more force).
  \item \textbf{Coordination of Variables}: In many tasks, corrections are needed to multiple state variables simultaneously. Representing this coordination through a single input may be desirable as it represents a coordinated behavior or latent dimension within the task. However, it may also be desirable to limit the number of state variables corresponding to a single user input to avoid confusing the user.
\end{itemize}

\subsection{Mapping Corrections to Low-Degree-of-Freedom Input}
A related question to what corrections a user should be able to make to the robot state is how to provide a user with an intuitive interface to provide these corrections. A correction system should consider:
\begin{itemize}[leftmargin=*]
    \item \textbf{Dimension of corrective input}: Depending on the task, there may be multiple corrections that are needed at any given time. Providing a single degree-of-freedom input offers simplicity that may make the device easier to use, but sacrifices expressiveness. We argue that this choice is often motivated by the task. If there is a clear single latent dimension of correction (e.g., abrasiveness while sanding) that is needed at a given time, perhaps this is sufficient. When a given task requires several types of corrections (i.e., multiple latent dimensions) at any given moment, a higher degree-of-freedom input may be necessary.
    \item \textbf{Spatial Grounding and Mobility}: Depending on the task, it may be desirable to give corrective input through a mobile device so that the user can interact with and assess the task during execution. For example, when sanding the operator may wish to closely inspect the surface to determine whether or not to reverse the execution. If the mobile input and necessary corrections are low-dimensional, it may be desirable to avoid spatial mapping. For example, a user could turn a mobile knob to apply more force without having to align the input with the direction of force. However, when an input device is stationary, it is logical to leverage spatial alignments for the corrections whenever possible. This becomes of increasing importance when multiple types of corrections may be needed.
    \item \textbf{Communication of Input mapping}: Depending on how the needed corrections vary over the task, there may be a need for different levels of communication about what corrections can be made during the execution. If the identified types of corrections are relatively consistent over the task duration, it may be sufficient to summarize the mapping and present it to the operator before the task (e.g., an engineer could explain the general mapping to an operator). Depending on the task, practice may also enable the operator to gain intuition about the mapping. However, if the correction ranges and variables vary dramatically over the execution, a real-time display may be needed such as an additional display or augmented reality (AR).
\end{itemize}

%% file: implementation.tex
\section{IMPLEMENTATION}
We designed an implementation of \emph{Corrective Shared Autonomy} which focuses on extracting correction variables, scaling, and mapping from expert demonstrations. Specifically, the implementation is tailored towards tasks involving interaction against known curved surfaces. In this section, we describe our design choices including (A) representing the nominal autonomous behavior and corrections based on natural demonstrations, (B) extracting correction components using Principal Component Analysis, (C) mapping human input to corrections, and (D) an override strategy for corrections outside of the range of the demonstrations.

\subsection{Data Preprocessing}
The original data consists of $N$ expert demonstrations. First, the demonstrations are aligned using dynamic time warping (DTW) \cite{dtw2009}. In addition to aligning the curves, this technique can also compute a timewarp between each of the demonstrations, which we subsequently use as an element of the demonstration state vector. In this work, we use the Euclidean distance of the Cartesian position as the distance metric for alignment. While other metrics leveraging more state variables could also be used, we chose Cartesian position to avoid combining different units.

We consider tasks that involve physical interaction with the environment. Movement in free-space (e.g., movements between passes of sanding) and movement against a surface (e.g., sanding) are commonly represented and controlled using different state variables. For example, the position of free-space motion consists of three Cartesian variables whereas the position during movement on a surface is represented with two surface coordinates. Thus, after the timewarping, the demonstration data is partitioned into segments consisting of free-space motion and motion against a surface (which are commanded using hybrid control). The segmentation is based on a filtered force threshold. The particular state variables for each segment vary depending on the task. The proposed method for extracting the correction variables and magnitudes is amenable to any choice of robot state variables or segmentation. Details of the state variables used in our experimental evaluation appear in Section \ref{sec:experiments}.

\subsection{Nominal Representation}
Encoding each segment of demonstration data as a robot behavior requires learning a nominal behavior that is most consistent with the multiple demonstrations. We choose to use one common learning framework, Dynamic Movement Primitives (DMPs) \cite{Ijspeert2013}, which offer benefits such as smoothness, stability, and trivial augmentation of the execution rate. A Dynamic Movement Primitive for a state vector, $\textbf{x}$, can be represented with the following set of dynamical systems:
\begin{gather}
    \tau\ddot{\textbf{x}}-\alpha(\beta(\textbf{g}-\textbf{x})-\dot{\textbf{x}})=\textbf{f}(s)\label{eq:dmp}\\
    \tau\dot{s}=-as
\end{gather}
where $s$ is the phase variable of the canonical system, $a$ is the related constant of the canonical system, $\tau$ is the time constant of the second-order dynamical system, $\textbf{g}$ is the goal state, $\alpha$ and $\beta$ are positive constants determining the roots of the dynamical system, and $\textbf{f}(s)$ is the nonlinear forcing function. The nonlinear forcing function is learned from the multiple demonstrations using locally weighted regression \cite{CalinonLee19}. During execution, the state values are determined by numerically integrating Equation \ref{eq:dmp}. Equation \ref{eq:dmp} assumes all state variables are independent, including the quaternions used for orientation (this is commonly addressed using a modified framework, CDMPs \cite{Morimoto2014}). Our implementation simply assumes the components change sufficiently slowly to be expressed independently and the final components are normalized to preserve the quaternion definition.

When later providing corrections to these behaviors, it is often desirable to be able to backtrack (i.e., reverse) the execution. For example, in this implementation, operators have a button that can be held to reverse the execution. However, the roots of the Dynamic Movement Primitive system become unstable when $\tau$ becomes negative. To address this, we learn forwards and backwards representations of Equation \ref{eq:dmp} for each segment and choose the corresponding system depending on the sign of the time constant.

\subsection{Correction State Variables and Scaling} \label{sec:scaling}
We believe that variance between expert demonstrations serves as a good indicator for corrections that may be needed for a given task. For each segment, we propose to use a per-frame (i.e., time sample) Principal Component Analysis (PCA) of the demonstration data.  Principal Component Analysis is a method for determining a set of orthonormal basis functions corresponding to the maximum directions of variance in a given set of data. The first principal component (PC), $\textbf{a}_{1}$, for a given time-sample is defined as:
\begin{equation}
\max_{\textbf{a}_{1}^T\textbf{a}_{1}=1} \sum_{i}\left( \left(\textbf{x}_{i}-\bar{\textbf{x}} \right) \cdot\textbf{a}_{1}\right)^2
\end{equation}
where $\textbf{x}_{i}$ is the state vector for demonstration $i$ and $\bar{\textbf{x}}$ is the mean state vector across demonstrations for the given time-sample. Subsequent principal components follow the same formulation subject to the additional constraint that they are orthonormal to any preceding components (i.e., for component, $k$, $\textbf{a}^T_{j}\textbf{a}_{k}=0,\: \forall j<k$). The principal components are computed using the singular value decomposition (SVD). The principal components of the state space correspond to combinations of the robot state variables that could potentially need corrections at a given time sample. By applying this algorithm at every sample and concatenating the results, each principal component ultimately yields a function of corrections over the execution.

When calculating the principal components, differences in state variable units can skew the components toward state variables with larger ranges. We therefore normalize each type of data by an expected range. Specific values used in our experimental tasks are given in \S\ref{sec:experiments}. We chose not to use the variance for each state for normalization, because noise in mostly static variables can skew the principal components. Additionally, the principal components represent an axis in the state space corresponding to the maximum variance, which is non-unique, and can flip in direction between samples. Therefore, we compute a dot product between successive samples and flip the direction when it is negative.

Finally, the principal components need to be transformed into a final scaled correction. The variance can be extracted by scaling the principal component by the corresponding singular value from the SVD. In this work, we first scale to three standard deviations (i.e., a scaling factor of $3\sigma_{i}N^{-1/2}$, where $\sigma_{i}$ is the singular value) and then to the original normalization factors to restore proper units. During execution, the final correction to each state variable, which is a combination of principal components (described in the next section), is filtered to prevent discontinuity.

\subsection{Human Input}\label{section:mapping}
Another challenge is how to allow a user to specify these coordinated corrections in an intuitive manner. In this section, we propose two potential approaches which are tailored for different distributions in the variance of the data. 

\subsubsection{One Degree of Freedom}
When the variance of the task can generally be described with a single principal component over the task execution, the user can provide corrections through a 1-DOF input. One benefit of this simple approach is that the first principal component, which describes the state-variable variance can be directly mapped to the device degree of freedom with little ambiguity in the mapping. The only uncertainty corresponds to how to map the input direction to the principal component direction (i.e., does moving the input in one direction increase or decrease the correction along the principal component). In this work, we assume that the input is not spatially grounded (e.g., mobile input). With a single degree-of-freedom, we argue this choice can often be arbitrary as long as it is consistent (i.e., people can learn to use either mapping). However, it may also be convenient to align the direction based on an important state variable. For example, in our experiment, the force represents the largest percentage of variance and thus, the principal component is aligned such that turning a rotational 1-DOF knob clockwise corresponds to more applied force.

\subsubsection{Three Degree of Freedom}
When corrections in multiple directions are needed, we propose to use a three degree-of-freedom device where the composition of the final correction from the principal components is dictated by the best-fit spatial alignment. We observe that most principal components contain at least one component that has a spatial correspondence. For example, in a drilling operation, the coordination of applied force, drill speed, and spray of cutting oil contains one component with a clear spatial direction: the direction of applied force. We propose a heuristic where the final corrective input is determined by assigning a spatial direction to each principal component based on the subset of state variables with a spatial correspondence (e.g., positions, forces -- defined as the set $spatial$) as summarized in Algorithm \ref{alg:spatial3dof}. The final correction is computed as a weighted sum (i.e., dot product) of the user input and the principal component directions.
\begin{algorithm}
\caption{PC Mapping for 3-DOF Input}\label{alg:spatial3dof}
\begin{algorithmic}
\For{$k \in 1,2,3$}\algorithmiccomment{\footnotesize\ttfamily{each principal component}}
 \State $\textbf{d}_{k}\gets \sum\limits_{l}\textbf{a}_{i,l} \mathbb{1}_{l \in \{\textrm{spatial}\}}$ \algorithmiccomment{\footnotesize\ttfamily{sum spatial variables}}
 \For{$r \in 1..(k-1)$}
    \State $\textbf{d}_{k}\gets \textbf{d}_{k}-proj(\textbf{d}_{k},\textbf{d}_{r})$ \algorithmiccomment{enforce orthogonality}
 \EndFor
 \State $\textbf{d}_{k} \gets \textbf{d}_{k}/||\textbf{d}_{k}||$ \algorithmiccomment{normalize}
\EndFor
\State $\delta \textbf{y} = \sum\limits_{k}\left(\textbf{u}\cdot \textbf{d}_{k}\right)\textbf{a}_{i}$ \algorithmiccomment{map user input to PCs}
\end{algorithmic}
\end{algorithm}
For example, a principal component could consist of of increased force, increased sanding speed, and a slower tangential velocity. If the spatial direction is assigned to be downward based on the force, when the users provides downward input, corrections will be apply simultaneously to all three variables.


\subsection{Range and Override}
While variance in expert demonstrations is intended to give a conservative range for corrections, there may be instances where the necessary correction is outside the range seen in the demonstrations (e.g., a defect that requires even more sanding force than the examples seen previously). To address, this we propose an override strategy that uses custom haptic profiles to partition the input space.

For context, all of the input devices operate in position regulation mode similar to that of a joystick, where in the absence of input, the device will remain in a centered, zero position. For each degree-of-freedom, the input is partitioned into two regions, which are enforced by haptic resistive forces that pull towards the zero position. The first region corresponds to corrections in the ranges of what was seen in the demonstrations and is enforced by a critically damped lower stiffness. Input in this region will map to a fixed amount of correction to the robot state variables. Once the user reaches to the edge of this region, there is a second, larger stiffness resembling a haptic wall that the user can penetrate in order to provide overrides. Input in this region uses an integral controller (i.e., the amount of corrections grows with user input that penetrates into the haptic wall). Because the amount of correction needed in the override range is unknown, integral control gives the flexibility to achieve arbitrarily-large levels. The override mapping and resistive force can be summarized as:
\begin{align}
    u_{t}=\left\{
\begin{array}{ll}
      d & |d|\leq d_{wall} \\
      u_{t-1}+\gamma\: sgn(d)\left( |d|-d_{wall} \right) & |d|>d_{wall} 
\end{array} 
\right.\\
    f_{t}=\left\{
\begin{array}{ll}
      kd+b\dot{d} & |d|\leq d_{wall} \\
      kd+k_{wall}(d-d_{wall})+b\dot{d} & |d|>d_{wall} 
\end{array} 
\right.
\end{align}
where $u_{t}$ is the user input at time $t$; $d$ is the device input (relative center position); $d_{wall}$ is magnitude of the wall distance for the proportional-integral control switch; $\gamma$ is a scaling factor for the integral control; $k$ is the lower stiffness for the non-override range; $b$ is a damping term for stability; and $k_{wall}$ is the higher stiffness. For multiple degrees of freedom, this strategy is implemented in each axis. Ideally, the axes would be aligned with the principal component directions, however, given that the principal components can evolve over the execution, this leads to inconsistent forces on the haptic device (e.g., an input which was proportional could become an override if the directions switch). In this work, we implement the override strategy in the device axes (i.e., the haptic wall resembles a cube), acknowledging this could lead to proportional inputs larger than $d_{wall}$ depending on the alignment of the device and principal component axes (e.g., the PC aligns with the diagonal of the cube).

%% file: experimentalevaluation.tex
\begin{figure}[]
\centering
\includegraphics[width=3.25in]{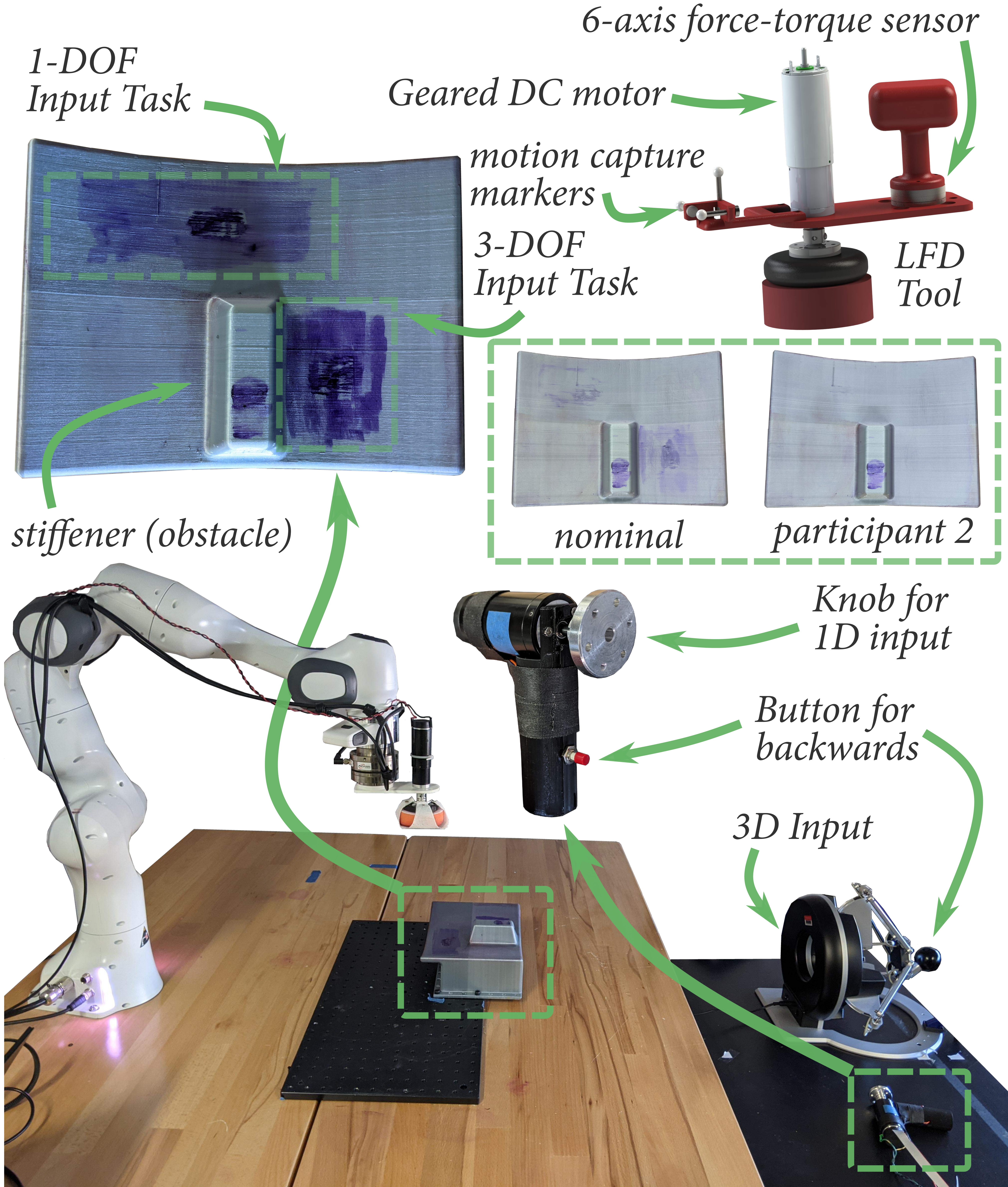}
\caption{Experimental tasks, instrumented LFD tool, and input methods.}
\label{fig:expsetup}
\vspace{-8pt}
\end{figure}

\section{EXPERIMENTAL RESULTS AND EVALUATION} \label{sec:experiments}
This section describes our experimental tasks, hardware setup, and results from the PCA-based method in \S\ref{sec:scaling} and a user study assessing the input mappings proposed in \S\ref{section:mapping}. Our study aimed to assess whether the PCA-based method could enable automated correction scaling and variable inference in Corrective Shared Autonomy.

\subsection{Task Description}
We designed two surface cleaning tasks which resemble release-agent cleaning of aerospace composites. The tasks consist of removing marker and paint from a surface. The marker required less effort to remove than the paint. The experimental setup and tasks are shown in Figure \ref{fig:expsetup}. Both tasks required cleaning a scaled replica of a fuselage barrel. The piece contains a stiffener that serves as an obstacle. The cleaning tool was a motorized polishing pad covered by a paper towel. The paper towel was soaked with alcohol to help with the removal. Expert demonstrations were performed by the authors to assure the task criteria were met. The demonstrations were used to encode the nominal behavior and allowable corrections for the user study in \S\ref{sec:userstudy}.

\subsubsection{Task 1: 1-DOF Corrections}
The first task was designed to contain a single dominant source of variation during the execution. The demonstrations consist of a single pass across the top section of the work piece. The path was consistent between demonstrations. Between trials, the amount and location of the paint was varied. When removing the paint, more force was applied, and the motor speed was increased. The learning used a total of six demonstrations.

\subsubsection{Task 2: 3-DOF Corrections}
The second task was designed to contain two dominant sources of variation during the execution. The demonstrations consist of a single pass of the tool next to the stiffener obstacle. During demonstrations, the closeness to the obstacle edge was varied. Additionally, the amount of force and sanding speed were varied to address hard-to-remove sections of paint, similar to the first task. A total of eight demonstrations were used for the learning.

 
\subsection{System Details}
Demonstration data was tracked using an Optitrack motion capture system and an ATI Mini40 6-axis force-torque sensor as shown in Figure \ref{fig:expsetup}. When using the hand tool, the speed of the rotating tool  was controlled using a Speedgoat Real-Time Target controller and set using a second handle with a potentiometer. Table \ref{table:statevar} shows the state variables and ranges of state variables (for the PCA normalization) that were recorded from the various demonstrations. The tool speed was the voltage command sent to velocity-control amplifier (0-5V). The geometries for hybrid control were represented using B-Spline Surfaces ($u$ and $v$ are the parameterized coordinates) \cite{Mortenson1990}. When interacting with the surface, the roll and pitch angles represented rotations relative to the surface normal and were defined with respect to the principal axes of the surface (e.g., $d\textbf{r}/du$ and $d\textbf{r}/dv$, where $\textbf{r}$ is the Cartesian position) rather than the velocity direction.


\begin{table}[b]
\centering
  \caption{State variables and ranges for normalization.}
  \label{table:statevar}
  \setlength\tabcolsep{4pt}
  
  \begin{tabular}{m{1.2cm}m{1.1cm}m{1.8cm}m{1.3cm}m{1.8cm}}
    \toprule
  \multicolumn{5}{c}{\textbf{Position Control}}\\
    \midrule
    \textbf{Description} & Position & Orientation & Tool Speed & Execution Rate \\
    \textbf{Variable(s)} & <$x,y,z$> & <$q_{x}$, $q_{y}$, $q_{z}$, $q_{w}$> & $v_{tool}$ & $\Delta n$ \\
    \textbf{Range} & 1 m & 1.0 & 5 V & 2 \\
    \end{tabular}
    \begin{tabular}{m{1.15cm}m{1.85cm}m{0.6cm}m{1.1cm}m{1.3cm}m{0.95cm}}
    \midrule
    \multicolumn{6}{c}{\textbf{Hybrid Control}}\\
    \midrule
    \textbf{Description} & Surface Coords. & Force & Tool Spd. & Exec. Rate & Roll/Pitch \\
    \textbf{Variable(s)} & <$u$,$v$> & $f_{n}$ & $v_{tool}$ & $\Delta n$ & <$\theta_{u},\theta_{v}$> \\
    \textbf{Range} & 1.0 & 20 N & 5 V & 2 & 1.57 rad \\
\bottomrule
  \end{tabular}
\end{table}

All robot executions used a Franka Emika Panda collaborative robot equipped with an ATI Axia80 6-axis force-torque sensor and the same geared DC motor end effector used in demonstrations. The robot was commanded in joint velocity using pseudo-inverse based inverse kinematics, and the hybrid control was implemented using an admittance model (i.e., reading forces, commanding velocity with a low proportional gain). To better facilitate contact with the environment, the robot was operated in joint impedance mode (400 N/rad). During execution, corrections to force and the execution rate were limited (i.e., saturated) to avoid faulting the robot. Depending on the task, user input was provided using either an ungeared Maxon DC motor mounted on a custom handle or a Force Dimension Omega 3 haptic device as shown in Figure \ref{fig:expsetup}. Each input device had a button that could be pressed and held to reverse the execution.

\begin{figure}[t]
\centering
\includegraphics[width=3.4in]{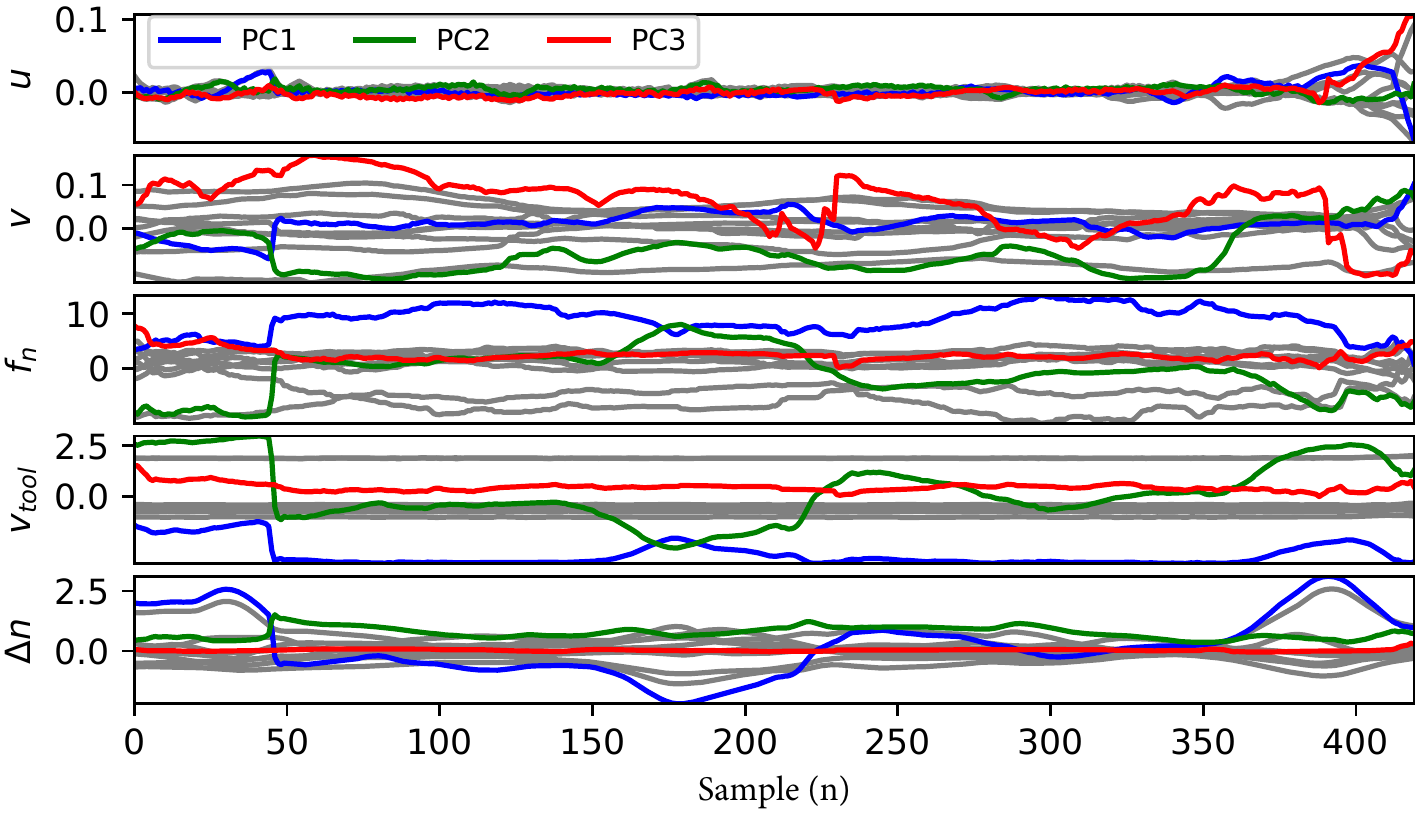}
\caption{Example of extracting corrections from demonstration from the surface interaction segment of Task 2. The mean-removed demonstrations are shown in gray in addition to the first three principal components (projected onto the original dimensions). $\theta_u$ and $\theta_v$ contained minimal variance and are omitted for clarity.
}
\label{fig:pcresults}
\vspace{-4pt}
\end{figure}

\subsection{Extracting Correction Variables and Scaling}
We validated the results of our per-frame PCA method to assure appropriate correction principal components were identified for each of the tasks. Each demonstration was correctly segmented via force thresholding into three sections: approaching the surface, the pass over the surface, and retracting from the surface. For the surface pass of the first task, the first principal component consisted for 86 percent of the variance (averaged across samples). In the second task, the first principal component consisted of 80 percent and the second principal component consisted of 15 percent of the variance. In these examples, the variance contained in the first principal component was similar between tasks, however, a simple solution for ambiguous cases is to allow operators to determine whether the higher degree-of-freedom input is required based on the contents of the principal component and task semantics.

The corrections extracted from the principal components matched the intentions of the given demonstrations. In Task 1, the first principal component combined an increased force, increased tool speed, and decreased tangential velocity. We present the results of the hybrid control segment of Task 2 as an example (see Figure \ref{fig:pcresults}). The first principal component was mainly a coordination of increased force and sanding speed. The dominant behavior in the second principal component was a correction to the second surface parameter, $v$, corresponding to motion closer to the obstacle. The third principal component, which accounted for four percent of the variance, also contained a correction to the second surface coordinate. Additionally, in the first 50 samples, the second principal component contained corrections to the applied force, which later appear in the first principal component. These splits across principal components suggest that rotating the principal components in the future (e.g., varimax) may be helpful to minimize the distribution of similar corrections across principal components for multiple degree-of-freedom mappings. Summarizing, our PCA method successfully extracted principal components that maximized variance and represented appropriate corrections for the experimental tasks. We plan to explore other extractions that further balance variance and consistency in the future.

\subsection{User Evaluation of Input Mappings}
\label{sec:userstudy}
To determine the feasibility of the input mappings in \S\ref{section:mapping}, we performed an initial validation study involving nine participants (six male, three female), aged 18--31 ($M=21.2$, $SD=3.8$), recruited from the UW--Madison campus. The procedure was administered under a
protocol approved by the Institutional Review Board
(IRB) of the UW--Madison.

\subsubsection{Procedure}
After providing written informed
consent, participants completed the two tasks presented in a counter-balanced order. Before each evaluation trial, participants were allowed to practice on a similar practice work piece. For each task, participants were provided with a summary of the input mapping (e.g., turning the knob applies more force and spins faster, pushing left or right modifies the path).

\begin{table}[t]
\centering
  \caption{Usability (SUS) and performance for each task. }
  \label{table:susandperf}
  \setlength\tabcolsep{4pt}
  \begin{tabular}{lll}
    \toprule
    \textbf{Task} & \textbf{Metric} & \textbf{Result (Mean (SD))}\\    
    \midrule
    \multirow{3}{*}{\textbf{1-DOF}} & SUS & 88.6 (9.9) \\
 & Color Removed & 0.90 (.07)\\ 
 & Color Removed - Nominal & 0.77 (0.05) \\
    \midrule
    \multirow{3}{*}{\textbf{3-DOF}} & SUS & 88.1 (6.6) \\
 & Color Removed & 0.96 (0.03)\\ 
 & Color Removed - Nominal & 0.57 (0.06) \\
\bottomrule
  \end{tabular}
\end{table}

\subsubsection{Metrics}
After each recorded trial, participants filled out the System Usability Scale (SUS) \cite{sus1995}. Additionally, images from before and after the tasks were processed to determine the percentage of marker and paint removed. The amount of color removed was estimated by calculating the standard deviation of the RGB color channels (the final color is gray with approximately zero standard deviation). The value was calculated as the difference in color between images divided by the starting amount of color, where each value was biased by a gray calibration value.

\subsubsection{Results}
All participants were able to complete both tasks. Two participants collided with the stiffener during practice, which consequently ripped the paper towel. However, no incidents occurred during the recorded trials. Participants utilized different strategies (see supplementary video). Some used a \textit{preemptive} strategy, applying corrections to force and tool speed preemptively as the tool passed over areas with large amounts of paint. Others used a \textit{posterior} strategy where corrections were applied after observing the result. In this case, participants reversed the execution and then increased force and tool speed. In all cases, participants backtracked the execution for at least one additional pass.

The SUS and performance results are reported in Table \ref{table:susandperf}. For both input mapping methods, the SUS scores were in the top quartile of system ratings (see Bangor et al. \cite{susadjective2008}). An example result of the nominal behavior and a participant trial can be seen in Figure \ref{fig:expsetup}. The nominal behavior aggregates multiple robot trials, where the location of paint was varied, which was insufficient to remove all paint from the surface (i.e., the average force and tool speed was less than required). Similar cleaning results to those shown in Figure \ref{fig:expsetup} were seen across all participants. Overall, visual inspection of the pieces and the color analysis results indicate that our method allowed participants to correct the nominal behavior to remove all visible paint on the surface.

%% file: conclusions.tex
\section{GENERAL DISCUSSION}

Our implementation successfully allowed users to provide corrections in a surface cleaning context. However, there were a number of limitations that serve as inspiration for future work. First, the demonstrations were limited to behaviors that could be learned through direct regression (e.g., demonstrations of the same task and environment that can be modeled with DMPs). The method also assumed that all necessary corrections would appear in the variance of the demonstrations and that the resulting principal components would have spatial correspondences that could be used for input mapping. In the future, we would like to explore more conservative overrides, input mappings for degenerate principal components (i.e., no spatial direction), corrections to state variables without variance in the demonstrations (e.g., a new collision in the path), and more advanced learning models for correction inference. We would also like to test the end-to-end system with expert technicians who can provide demonstrations and assess the resulting \emph{Corrective Shared Autonomy}. In this work, the types of corrections that could be made were consistent over the execution and were summarized to participants before training. We believe that further studying effective communication of allowable corrections to the operator is a crucial piece of designing an effective system, particularly if the corrections vary over the course of the execution.
